\documentclass[sigconf, 10pt, nonacm]{acmart}
\usepackage{hyperref}       
\usepackage{url}            
\usepackage{booktabs}       
\usepackage{amsfonts}       
\usepackage{nicefrac}       
\usepackage{microtype}      
\usepackage{xcolor}         
\usepackage{amsmath}
\usepackage{algorithm}
\usepackage{algorithmic}
\usepackage{enumitem}
\usepackage{colortbl}
\usepackage{array}
\usepackage{geometry}
\usepackage{graphicx}
\usepackage{caption}
\usepackage{subcaption}
\usepackage{svg}
\usepackage{longtable}
\usepackage{arydshln}
\usepackage{multirow}
\usepackage{booktabs}
\usepackage{caption}
\newtheorem{theorem}{Theorem}[section]

\begin{document}

\title{Enhancing Parameter Efficiency and Generalization in Large-Scale Models: A Regularized and Masked Low-Rank Adaptation Approach}

\author{Yuzhu Mao}
\email{myz20@mails.tsinghua.edu.cn}
\affiliation{%
  \institution{\footnotesize Tsinghua-Berkeley Shenzhen Institute \\Tsinghua Shenzhen International Graduate School \\Tsinghua University}
  \country{\footnotesize China}
}

\author{Siqi Ping}
\email{psq22@mails.tsinghua.edu.cn}
\affiliation{%
  \institution{\footnotesize Tsinghua-Berkeley Shenzhen Institute \\Tsinghua Shenzhen International Graduate School \\Tsinghua University}
  \country{\footnotesize China}
}

\author{Zihao Zhao}
\email{zhao-zh21@mails.tsinghua.edu.cn}
\affiliation{%
  \institution{\footnotesize Tsinghua-Berkeley Shenzhen Institute \\Tsinghua Shenzhen International Graduate School \\Tsinghua University}
  \country{\footnotesize China}
}

\author{Yang Liu}
\email{liuy03@air.tsinghua.edu.cn}
\affiliation{%
  \institution{\footnotesize Institute for AI Industry Research (AIR) \\ Tsinghua University \\ Shanghai AI Lab}
\country{\footnotesize China}
}

\author{Wenbo Ding}
\email{ding.wenbo@sz.tsinghua.edu.cn}
\affiliation{%
  \institution{\footnotesize Tsinghua-Berkeley Shenzhen Institute \\Tsinghua Shenzhen International Graduate School \\Tsinghua University \\ Shanghai AI Lab}
\country{\footnotesize China}
}
\authornote{Corresponding Author}

\renewcommand{\shortauthors}{Mao et al.}
\acmArticleType{Research}
\acmCodeLink{https://github.com/borisveytsman/acmart}
\acmDataLink{htps://zenodo.org/link}
\acmContributions{BT and GKMT designed the study; LT, VB, and AP
  conducted the experiments, BR, HC, CP and JS analyzed the results,
  JPK developed analytical predictions, all authors participated in
  writing the manuscript.}
\begin{abstract}
  Large pre-trained models, such as large language models (LLMs), present significant resource challenges for fine-tuning due to their extensive parameter sizes, especially for applications in mobile systems. To address this, Low-Rank Adaptation (LoRA) has been developed to reduce resource consumption while maintaining satisfactory fine-tuning results. Despite its effectiveness, the original LoRA method faces challenges of suboptimal performance and overfitting. This paper investigates the intrinsic dimension of the matrix updates approximated by the LoRA method and reveals the performance benefits of increasing this intrinsic dimension. By employing regularization and a gradient masking method that encourages higher intrinsic dimension, the proposed method, termed \textbf{R}egularized and \textbf{M}asked LoRA (RM-LoRA), achieves superior generalization performance with the same or lower trainable parameter budget compared to the original LoRA and its latest variants across various open-source vision and language datasets.    
\end{abstract}


\maketitle

\vspace{-0.25cm}
\section{Introduction}

Large pre-trained models, like large-scale language models (LLMs), have showcased remarkable performance across a variety of tasks in computer vision and natural language processing \cite{zhang2022opt,brown2020language,touvron2023llama,ouyang2022training}. Nonetheless, fine-tuning these models for specific downstream tasks often presents substantial resource challenges due to their extensive parameter sizes. In this context, parameter-efficient fine-tuning (PEFT) methods have been extensively explored to alleviate resource consumption while preserving or enhancing fine-tuned model performance \cite{zaken2022bitfit,hu2021lora,li2021prefix,lester2021power,vu2022spot,guo2021parameter,zhang2023llama,liu2022few,houlsby2019parameter,liu2022p,liu2023gpt,sung2021training,sung2022lst,mao2022unipelt,leemixout}. Among these methods, Low-Rank Adaptation (LoRA), which involves freezing the pre-trained weights and approximating updates in weight matrices using the multiplication of two low-rank matrices, has emerged as a promising approach to balance computational efficiency and task performance during the fine-tuning process of large models \cite{hu2021lora}.

Despite its effectiveness, LoRA fine-tuning encounters challenges in determining the optimal size of LoRA matrices for a given model and task. On one hand, excessively small matrices, with limited number of trainable parameters, inevitably harm training convergence and generalization performance. On the other hand, large matrices introduce redundant trainable parameters, which could be reduced to enhance parameter efficiency. Moreover, some studies have indicated that large LoRA matrices may exacerbate overfitting, as redundant parameters primarily contribute to training accuracy rather than test accuracy \cite{qiang2024bilora, karimi2021compacter}.

Several approaches have been proposed to determine or adaptively adjust the size of LoRA matrices, often referred to as the LoRA rank $R$, for improved efficiency and generalization. However, none of these methods investigate the intrinsic dimension $r$ of the approximated matrix update $\Delta \mathbf{W} = \mathbf{B}\mathbf{A}$ given by the product of low-rank LoRA matrices $\mathbf{A}$ and $\mathbf{B}$. This intrinsic dimension $r$, instead of the previously studied LoRA rank $R$, has been proven to play a crucial role in LoRA fine-tuning. Specifically, \citet{zeng2023expressive} theoretically demonstrated that for fully connected neural networks, the LoRA approximation error given by an approximated update $\Delta \mathbf{W} = \mathbf{B}\mathbf{A}$ with intrinsic dimension $r$ is related to the $r$-th singular value of the discrepancy $\mathbf{E} = \mathbf{W}_{\text{target}} - \mathbf{W}_{\text{frozen}}$ between the target weight matrix and the frozen pre-trained weight matrix. 

In other words, encouraging the intrinsic dimension $r$ of $\Delta \mathbf{W}$ to approximate the given LoRA rank $R$ benefits the generalization of LoRA fine-tuning under a given trainable parameter budget. Inspired by this theoretical conclusion, this paper first adopts a regularization technique to encourage LoRA matrices to span a higher intrinsic rank in their parameter space. Additionally, to maintain a reasonable budget of trainable parameters, a gradient masking method is introduced to randomly mask a subset of parameters in each epoch instead of updating all parameters in LoRA matrices. Experiments on multiple datasets have proven this method also helps promote the growth of the intrinsic rank $r$ and thus yields lower approximation error and better generalization performance.

The contributions of this paper can be summarized as follows: 1) This paper extends previous theoretical bounds for LoRA approximation error from simulated datasets to real-world datasets, providing further insights into the trade-off between LoRA rank $R$ and generalization performance.
2) Based on the analysis of LoRA rank and generalization performance, this paper designs a strategy for fine-tuning LoRA matrices that encourages the growth of intrinsic rank $r = \text{rank}(\Delta \mathbf{W})$ within the LoRA parameter space defined by $R$. This strategy effectively alleviates the problem of overfitting the training data by encouraging the LoRA matrices to explore the parameter space. 3) The experimental results across multiple open-source datasets demonstrate that this \textbf{R}egularized and \textbf{M}asked version of LoRA (\textbf{RM-LoRA}) method manages to strike a better efficiency-generalization tradeoff compared to the original LoRA method and its state-of-the-art variations, with better generalization performance achieved with the same or lower trainable parameters budget.

\vspace{-0.26cm}
\section{Related Works}
In attempts to address the computational challenges posed by updating the enormous amount of weights in large pre-trained models, LoRA, proposed by \citet{hu2021lora}, achieves outstanding model generalization with a significantly reduced budget of trainable parameters during fine-tuning. However, LoRA still faces the challenges of sub-optimal performance and overfitting. Previous research addressing LoRA's main challenges is briefly discussed as follows:


\textbf{Sub-Optimal Performance}. While LoRA has demonstrated remarkable parameter efficiency and generalization performance, it can lead to suboptimal fine-tuning of large-scale models with high embedding dimensions \cite{hayou2024lora+}. In some cases, there is a contradictory phenomenon where a higher LoRA rank doesn't necessarily bring better performance than a lower LoRA rank. Much research has been devoted to further balancing the efficiency and performance achieved by LoRA, including the adaptive choice of LoRA rank \cite{zhang2023adaptive,ding2023sparse,valipour2023dylora}, adjustment of learning rate \cite{hayou2024lora+}, random projection \cite{kopiczko2023vera}, derivative-free optimization \cite{jin2024derivative}, and pre-trained weights optimization \cite{zi2023delta}. Nevertheless, none of these methods consider the role of LoRA updates' intrinsic dimension in mitigating the performance gap under a given LoRA rank setting.

\textbf{Overfitting.} Fine-tuning large pre-trained models with a large number of parameters can easily encounter overfitting, resulting in suboptimal generalization, especially on test data \cite{karimi2021compacter}. In the AdaLoRA method, the LoRA matrices for less important pre-trained weight matrices are assigned a lower rank to prevent overfitting \cite{zhang2023adaptive}. However, according to the experiments of \citet{qiang2024bilora}, LoRA and AdaLoRA still clearly overfit the training data as fine-tuning advances, with decreases in training losses but increases in test losses. To alleviate the overfitting problem, \citet{qiang2024bilora} developed the BiLoRA method, which iteratively trains different subsets of trainable parameters using different subsets of training data. Furthermore, \citet{hayou2024lora+} pointed out that the suboptimality of LoRA and its variants is due to some directions of LoRA matrices not being sufficiently updated, and thus the change in model weights approximated by LoRA is restricted by the vector (sub)space generated by the LoRA matrices' columns at initialization.

The previous analysis of LoRA's existing limitations and solutions gives rise to the idea and method employed in this work. Specifically, for better performance under a given LoRA rank setting, this paper proposes a fine-tuning strategy that promotes the growth of the intrinsic dimension of LoRA updates through regularization and gradient masking, bridging the gap between practical performance and theoretical optimal performance. 

\section{RM-LoRA Method}
\label{sec:method}
\subsection{Preliminary}
\textbf{Transformer Models.} A transformer-based pre-trained model typically involves $L$ stacked encoder/decoder blocks, with a multi-head attention module followed by a fully connected feed-forward network (FFN) in each block. Given an input sequence $\mathbf{X} \in \mathbb{R}^{n \times d}$, the output of the multi-head attention module can be written as:
\begin{align}
    &\mathrm{MultiHead}(\mathbf{X}) = \mathrm{Concat}(\mathrm{head}_1, \ldots, \mathrm{head}_h)\mathbf{W}^O, \nonumber \\
    &\text{with}~\mathrm{head}_i = \mathrm{Attention}(\mathbf{Q}_i, \mathbf{K}_i, \mathbf{V}_i), \nonumber \\
    &\text{and}~ \mathrm{Attention}(\mathbf{Q}_i, \mathbf{K}_i, \mathbf{V}_i) = \mathrm{softmax}\left(\frac{\mathbf{Q}_i \mathbf{K}_i^T}{\sqrt{d_k}}\right)\mathbf{V}_i, \label{eq:multihead_attention}
\end{align}
where $\mathbf{Q}_i = \mathbf{X} \mathbf{W}^Q_i$, $\mathbf{K}_i = \mathbf{X} \mathbf{W}^K_i$, and $\mathbf{V}_i = \mathbf{X} \mathbf{W}^V_i$ are matrices of queries, keys, and values of $\mathrm{head}_i$ respectively, with projection matrices $\mathbf{W}^Q_i \in \mathbb{R}^{d \times d_k}$, $\mathbf{W}^K_i \in \mathbb{R}^{d \times d_k}$, $\mathbf{W}^V_i \in \mathbb{R}^{d \times d_v}$, and $\mathbf{W}^O \in \mathbb{R}^{hd_v \times d}$. We refer readers to \cite{vaswani2017attention} for a more comprehensive introduction to attention calculations in general.

Given the output of the multi-head attention module, the FFN further projects the $d$-dimensional output $\mathbf{X}'$ for each position. A two-layer FFN with a ReLU activation operates as follows:
\begin{equation}
   \mathrm{FFN}(\mathbf{X}') = \max(0, \mathbf{X}' \mathbf{W}_{f_1} + b_1) \mathbf{W}_{f_2} + b_2,
\end{equation}
where $\mathbf{W}_{f_1} \in \mathbb{R}^{d \times d_m}$ and $\mathbf{W}_{f_2} \in \mathbb{R}^{d_m \times d}$. Moreover, a residual connection followed by layer normalization is applied to each layer to generate the output of each transformer block given the input sequence $\mathbf{X}$ in the following way:
\begin{equation}
   \mathrm{LayerNorm}(\mathbf{X} + \mathrm{FFN}(\mathrm{LayerNorm}(\mathbf{X} + \mathrm{MultiHead}(\mathbf{X})))).
\end{equation}

\textbf{LoRA Fine-tuning.} For a pre-trained matrix $\mathbf{W}_0 \in \mathbb{R}^{d_1 \times d_2}$, LoRA, as proposed by \citet{hu2021lora}, approximates its update $\Delta \mathbf{W}$ by $\Delta \mathbf{W} = \mathbf{B}\mathbf{A}$, where $\mathbf{A} \in \mathbb{R}^{R \times d_2}$ and $\mathbf{B} \in \mathbb{R}^{d_1 \times R}$ with rank $R \ll \min(d_1, d_2)$ . During model fine-tuning, the weight matrix $\mathbf{W}_0$ is frozen, with only the LoRA adapters $\mathbf{A}$ and $\mathbf{B}$ being trainable. The modified LoRA forward pass is:
\begin{align}\label{eq:forward_lora}
    \mathbf{h} = \mathbf{W}_0 \mathbf{x} + \Delta \mathbf{W} \mathbf{x} = \mathbf{W}_0 \mathbf{x} + \mathbf{B}\mathbf{A}\mathbf{x}.
\end{align}

Typically, the low-rank matrice $\mathbf{A}$ is initialized using a random Gaussian distribution, while $\mathbf{B}$ is initialized to zero, ensuring $\Delta \mathbf{W} = 0$ at the start of fine-tuning. Current approaches to fine-tuning large pre-trained models with LoRA apply a pair of matrices to all weight matrices involved in each transformer block's multi-head attention module and FFN \cite{he2021towards, zhang2023adaptive}.

\textbf{The Expressive Power of LoRA.} \citet{zeng2023expressive} investigates the LoRA approximation error under a mild non-singularity assumption. To begin with, a $L$-layer width-$D$ fully connected ReLU neural network is denoted as $\text{FNN}_{L,D}(\cdot; (\mathbf{W}_l)_{l=1}^L, (\mathbf{b}_l)_{l=1}^L)$, where $\mathbf{W}_l \in \mathbb{R}^{D \times D}$ are the weight matrices and $\mathbf{b}_l \in \mathbb{R}^{D}$ are the biases for each layer $l \in [L]$. In LoRA method, the primary objective is to adapt a pre-trained frozen FNN $f_0$ to approximate a target FNN $\bar{f}$, which are represented as follows:
\begin{align}
\text{Target FNN} ~ \bar{f} &:= \text{FNN}_{\bar{L},D}(\cdot; (\mathbf{\bar{W}}_l)_{l=1}^L, (\mathbf{\bar{b}}_l)_{l=1}^L), \\
\text{Frozen FNN} ~ f_0 &:= \text{FNN}_{L,D}(\cdot; (\mathbf{W}_l)_{l=1}^L, (\mathbf{b}_l)_{l=1}^L),
\end{align}
where $\mathbf{\bar{W}}_l \in \mathbb{R}^{D \times D}$ and $\mathbf{\bar{b}}_l \in \mathbb{R}^{D}$ are the weight matrix and bias vector for the $l$-th layer of the target model $\bar{f}$, while $\mathbf{W}_l \in \mathbb{R}^{D \times D}$ and $\mathbf{b}_l \in \mathbb{R}^D$ are those for the $l$-th layer of the pre-trained frozen model $f_0$.

With a LoRA rank setting $R \in [D]$, the frozen FNN $f_0$ is adapted into a new model $f$:
\begin{align}
\text{Adapted FNN} \: & f := \text{FNN}_{L,D} (\cdot ; (\mathbf{W}_l + \mathbf{\Delta W}_l)_{l=1}^L, (\mathbf{\hat{b}}_l)_{l=1}^L),
\end{align}
where $\Delta \mathbf{W}_l \in \mathbb{R}^{D \times D}$ is the weight update approximated by LoRA with $\text{rank}(\Delta \mathbf{W}_l) \leq R_l$, and $\mathbf{\hat{b}}_l$ is the updated bias vector or $l \in [L]$. Given that large pre-trained model tends to be overparameterized, it is reasonable to assume that $L \geq \bar{L}$, which means the pre-trained model is much deeper than the traget model to be approximated. Therefore, \citet{zeng2023expressive} further introduces an ordered partition $\mathcal{P} = \{\mathcal{P}_1, \ldots, \mathcal{P}_{\bar{L}}\}$ to partition the $L$ layers in the adapted model $f$, such that $\bigcup_{i=1}^{\bar{L}} \mathcal{P}_i = [L]$. Each partition element $\mathcal{P}_i \in \mathcal{P}$ consists of consecutive integers $l \in \mathcal{P}_i$, which indicate the index of layers in the adapted model that will be used to approximate the $i$-th layer in the target model.

With the above definition, the following theoretical result provides an upper bound on the approximation error for the adapted model. 

\begin{theorem}[Theorem 6 in \cite{zeng2023expressive}]
If \; $\sum_{l\in \mathcal{P}_i} R_l  \geq \text{rank} (\bar{\mathbf{W}}_i - \prod_{l\in \mathcal{P}_i} \mathbf{W}_l)$ for all $i \in [\hat{L}]$,  there exists LoRA adapters $(\Delta \mathbf{W}_l)_{l=1}^L$ with $\text{rank}(\Delta \mathbf{W}_l) \leq R_l$ and biases $(\mathbf{\hat{b}}_l)_{l=1}^L$ such that the adapted model $f$ can exactly approximate the target model $\hat{f}$.

Furthermore, define the approximation error of the $i$-th layer as $e_i =  \sigma_{\sum_{l\in \mathcal{P}_i} R_l +1} (\bar{\mathbf{W}}_i - \prod_{l\in \mathcal{P}_i} \mathbf{W}_l) $, and the magnitude of the weight parameters and the input as
\[
\beta := \max_{i\in [\bar{L}]}\left(\sqrt{\left\| \mathbf{\Sigma} \right\|_F} \prod_{j=1}^{i} \left\| \bar{\mathbf{W}}_{j} \right\|_F + \sum_{j=1}^{i} \prod_{k=j+1}^{i-1} \left\| \bar{\mathbf{W}}_{k} \right\|_F \left\| \bar{\mathbf{b}}_{j} \right\|_2 \right) \vee \sqrt{\left\| \mathbf{\Sigma} \right\|_F}.
\]

Then, there exists LoRA adapters $(\Delta \mathbf{W}_l)_{l=1}^L$ with $\text{rank}(\Delta \mathbf{W}_l) \leq R_l$ and biases $(\hat{\mathbf{b}}_l)_{l=1}^L$ such that for any input $\mathbf{x}$ with $\mathbb{E} \mathbf{x} \mathbf{x}^{T} = \mathbf{\Sigma}$, the approximation  error can be bounded as
\begin{equation}
    \mathbb{E} \left\| f(\mathbf{x}) - \bar{f}(\mathbf{x}) \right\|_2 \leq \beta \sum_{i=1}^{\bar{L}} \max_{k \in [\bar{L}]} \left( \left\| \bar{\mathbf{W}}_k \right\|_F + e_k \right)^{\bar{L}-i} e_i .
\label{eq:LoRA error}
\end{equation}
\end{theorem}

In the above bound, $\beta$ and $\left\| \bar{\mathbf{W}}_k \right\|_F$ capture the magnitude of the weight parameters in the target model and the input. The LoRA rank setting $R_l$ for all layers $l \in [L]$ in the adapted model contributes to this bound through the term $e_i$ for all $i \in [\bar{L}]$. The following section focuses on the interconnection among the constituting parts of the $e_i$ term and explains how this theoretical conclusion can be utilized to enhance LoRA adaptation on real-world datasets.

\subsection{Influence of the Intrinsic Dimension of LoRA Adapter $\Delta \mathbf{W}$}

Note that the partition $\mathcal{P}_i$ of the pre-trained model for the $i$-th layer in the target model is an intrinsic but unknown property during adaptation, and consequently, the number and index of layers $l \in \mathcal{P}_i$ are also unknown. Nevertheless, with a pre-trained model $f_0$ to be adapted and the target model $\bar{f}$ determined by a given downstream task, the partition can be considered deterministic, as can the discrepancy between the pre-trained model and the target model $\mathbf{E}_i = \bar{\mathbf{W}}_i - \prod_{l \in \mathcal{P}_i} \mathbf{W}_l$. Consider the case where the LoRA rank setting for each layer $l \in \mathcal{P}_i$ is the same as $R$. The $e_i$ term in Equation~\ref{eq:LoRA error} can be rewritten as:
\begin{equation}
    e_{i, \text{rank}(\Delta \mathbf{W}_{l \in \mathcal{P}_i}) \leq R} = \sigma_{\sum_{l \in \mathcal{P}_i} R +1} (\mathbf{E}_i) \; \textit{for each layer} \; i \in [\bar{L}],
\end{equation}
where $\text{rank}(\Delta \mathbf{W}_{l \in \mathcal{P}_i)}\leq R$ represents the LoRA adapter for each layer $l \in \mathcal{P}_i$ satisfies the rank constraint $\text{rank}(\Delta \mathbf{W}_l) \leq R$.

Clearly, increasing $\text{rank}(\Delta \mathbf{W}_l)$ helps relax the constraint on LoRA rank $R$ to achieve a certain level of approximation error. For example, consider two LoRA rank settings $R_1$ and $R_2$ with $R_1 < R_2$. If $\text{rank}(\Delta \mathbf{W}_l) \leq R_1 < R_2$, then $e_{i, \text{rank}(\Delta \mathbf{W}_{l \in \mathcal{P}_i}) \leq R_2}$ degenerates to $e_{i, \text{rank}(\Delta \mathbf{W}_{l \in \mathcal{P}_i}) \leq R_1}$ for $i \in [\bar{L}]$, despite the larger size of LoRA matrices under the LoRA setting of $R_2$. In this case, LoRA rank $R_1$ and $R_2$ yield the same LoRA approximation error $\mathbb{E} \| f(\mathbf{x}) - \bar{f}(\mathbf{x}) \|_2$ according to Equation~\ref{eq:LoRA error}.

\subsection{Regularization on LoRA Weights}

Let a pair of LoRA low-rank matrices be denoted as $\mathbf{W}^A$ and $\mathbf{W}^B$, respectively. To enforce the growth in rank of $\Delta \mathbf{W} = \mathbf{W}^B \mathbf{W}^A$, the following regularizer is first used to encourage $\mathbf{W}^A$ and $\mathbf{W}^B$ to be orthogonal:
\begin{equation}\label{eq:regularization}
\text{Reg}(\mathbf{W}^A, \mathbf{W}^B) = \| \mathbf{W}^A (\mathbf{W}^A)^{\top} - \mathbf{I} \|_F^2 + \| (\mathbf{W}^B)^{\top} \mathbf{W}^B - \mathbf{I} \|_F^2 .
\end{equation}
The orthogonality of $\mathbf{W}^A$ and $\mathbf{W}^B$ helps increase the $\text{rank}(\mathbf{W}^A)$ and $\text{rank}(\mathbf{W}^B)$. According to the lower bound for the rank of the matrix product, for matrices $\mathbf{A} \in \mathbb{R}^{R \times d_2}$ and $\mathbf{B} \in \mathbb{R}^{d_1 \times R}$, the rank of their product matrix $\mathbf{C} = \mathbf{B}\mathbf{A}$ satisfies $\text{rank}(\mathbf{C}) \geq \max(\text{rank}(\mathbf{A}) + \text{rank}(\mathbf{B}) - R, 0)$. This lower bound ensures the growth of the intrinsic rank of the LoRA adapter $\Delta \mathbf{W} = \mathbf{W}^A \mathbf{W}^B$ as the $\text{rank}(\mathbf{W}^A)$ and $\text{rank}(\mathbf{W}^B)$ increase with the regularizer shown in Equation~\ref{eq:regularization}. Note that there exist other alternative regularizers that theoretically can also encourage the growth of $\text{rank}(\Delta \mathbf{W})$, but are infeasible in reality due to considerations of differentiability, numerical stability, and computational costs\footnote{The nuclear (trace) norm involves expensive computation for singular value decomposition, especially with large matrices. Regularization on the determinants suffers from numerical instability since the determinant calculation is highly sensitive to small changes in the matrix elements. Constraints on eigenvalues or singular values of the matrix are not directly differentiable.}. 

\vspace{-0.1cm}
\begin{algorithm}
    \caption{Gradient Masking Algorithm}
    \label{alg:gradient_masking}
    \begin{algorithmic}[1]
        \STATE {\bfseries Input:} Total steps $T$, LoRA rank $R$, number of updated directions $\hat{r}$.
        \FOR{$t = 0$ to $T-1$}
            \FOR{each pair of LoRA weight matrices $(\mathbf{W}^A_t, \mathbf{W}^B_t)$ in the model}
            \STATE Sample a mini-batch data $\xi_t$ and compute the gradients $(\nabla_{\xi_t} \mathbf{W}^A_t, \nabla_{\xi_t} \mathbf{W}^B_t)$;
            \STATE Initialize gradient masks $(\mathbf{M}^A_t, \mathbf{M}^B_t) \leftarrow \mathbf{0}$ with the same shape as$(\nabla_{\xi_t} \mathbf{W}^A_t, \nabla_{\xi_t} \mathbf{W}^B_t)$;
             \STATE Construct the set $\mathcal{R}_t$ by randomly selecting $\hat{r}$ distinct integers from $\{1, 2, \ldots, R\}$.
             \FOR{each $i$ in $\mathcal{R}_t$}
             \STATE $\mathbf{M}^A_{t}[i, j] = 1$ for all $j = 1, 2, \ldots, R$.
             \ENDFOR
             \FOR{each $j$ in $\mathcal{R}_t$}
             \STATE $\mathbf{M}^B_{t}[i, j] = 1$ for all $i = 1, 2, \ldots, R$.
             \ENDFOR
             \STATE Apply gradient mask $\nabla_{\xi_t} \mathbf{W}^A_t \leftarrow \nabla_{\xi_t} \mathbf{W}^A_t \odot \mathbf{M}^A_t$, $\nabla_{\xi_t} \mathbf{W}^B_t \leftarrow \nabla_{\xi_t} \mathbf{W}^B_t \odot \mathbf{M}^B_t$.
             \STATE Perform optimization step $\mathbf{W}^A_t = \mathbf{W}^A_t - \eta \nabla_{\xi_t} \mathbf{W}^A_t$, $\mathbf{W}^B_t = \mathbf{W}^B_t - \eta \nabla_{\xi_t} \mathbf{W}^B_t$.
            \ENDFOR
        \ENDFOR
        \STATE \textbf{Output:} Updated LoRA weight matrices $(\mathbf{W}^A_T, \mathbf{W}^B_T)$ for each fine-tuned module.
    \end{algorithmic}
\end{algorithm}

\subsection{Gradient Masking for Partial Updates}

The gradient masking algorithm in RM-LoRA is designed to perform partial updates in LoRA matrices. The algorithm takes as input the total number of steps $T$, the LoRA rank $R$, and the number of directions $\hat{r}$ to update in each step. In each training step $t$, it samples a mini-batch of data $\xi_t$ and computes the gradients $\nabla_{\xi_t} \mathbf{W}^A_t$ and $\nabla_{\xi_t} \mathbf{W}^B_t$ for each pair of LoRA weight matrices. The corresponding gradient masks are first initialized to zero, before a set $\mathcal{R}_t$ of $\hat{r}$ distinct directions is randomly selected. The RM-LoRA method then sets the relevant entries in the gradient masks to one according to the selected directions. These masks are applied to the gradients to restrict the update directions. Finally, the algorithm updates the weight matrices $\mathbf{W}^A_t$ and $\mathbf{W}^B_t$ using the masked gradients, thus achieving the partial update of LoRA weight matrices. The complete process of gradient masking is summarized in Algorithm~\ref{alg:gradient_masking}.

\section{Experiments}
\subsection{Experimental Setup}
The details of the experiment for the evaluation of the proposed RM-LoRA method are outlined as follows: 

\textbf{Models and Datasets}. This paper compares our proposed RM-LoRA with the original LoRA and its recent variants across both computer vision and natural language tasks. For vision task, a Vision Transformer (ViT) model \cite{dosovitskiy2020image} is fine-tuned on the CIFAR-100 dataset. For language tasks, a DeBERTaV3 model \cite{he2022debertav3} is fine-tuned on the General Language Understanding Evaluation (GLUE) benchmark for language understanding \cite{wang2018glue} and Stanford Question Answering Dataset (SQuAD 1.1) for question answering \cite{squad1}.

\textbf{Baselines}. The following baselines are implemented within the same HuggingFace's Transformers framework \cite{wolf2019huggingface}. LoRA and its variants are all implemented using the LoRA public code-base\footnote{\url{https://github.com/microsoft/LoRA}} for fair comparison:
\begin{itemize}[leftmargin=0.5cm]
    \item \textit{Full fine-tuning} (FT) uses the pre-trained model as the initialization point and updates all parameters in the model through gradient backpropagation.
    \item \textit{LoRA} \citep{hu2021lora} approximates the incremental updates in pre-trained model weights by using the product of two trainable matrices with rank $R$.
    \item \textit{AdaLoRA} \citep{zhang2023adaptive} uses the product of three small matrices in the form of singular value decomposition to parameterize the updates in pre-trained model weights, and then prunes the singular values of lower importance in the diagonal matrix to achieve a pre-set total parameter budget $b$ across all adapter weight matrices. 
    \item \textit{SoRA} \citep{ding2023sparse} parameterizes the updates in pre-trained model weights similarly to AdaLoRA, with an additional gate unit in between, and controls the sparsity of the gate by pruning components with absolute values lower than a pre-set threshold $\lambda$.
\end{itemize}
For a fair comparison of all the LoRA variants, including our proposed RM-LoRA method, their performance is evaluated under the \textbf{same parameter budget during inference} in the following part of this paper.

\begin{table*}
\begin{center}
\caption{Results with DeBERTaV3 model on GLUE benchmark.}
\vspace{-0.2cm}
\renewcommand{\arraystretch}{1.5}
\label{tab:glue results}
\scriptsize
\begin{tabular}{c|c|ccccccccc}
\hline
\textbf{Method} & \textbf{\# T / I-Params} & \begin{tabular}[c]{@{}c@{}}\textbf{CoLA}\\ Mcc\end{tabular} & \begin{tabular}[c]{@{}c@{}}\textbf{STS-B}\\ Pearson / Spearman\end{tabular} & \begin{tabular}[c]{@{}c@{}}\textbf{QNLI}\\ Acc\end{tabular} & \begin{tabular}[c]{@{}c@{}}\textbf{MNLI}\\ Acc \end{tabular} & \begin{tabular}[c]{@{}c@{}}\textbf{WNLI}\\ Acc\end{tabular} & \begin{tabular}[c]{@{}c@{}}\textbf{RTE}\\ Acc\end{tabular} & \begin{tabular}[c]{@{}c@{}}\textbf{MRPC}\\ Acc / F1\end{tabular} & \begin{tabular}[c]{@{}c@{}}\textbf{QQP}\\ Acc / F1\end{tabular} & \begin{tabular}[c]{@{}c@{}}\textbf{SST-2}\\ Acc\end{tabular} \\ \hline
Full FT         & 184M / 184M              & 0.6676                                                    & 0.8919 / 0.8902                                               & 0.9396                                                    & 0.9066                                                    & 0.563                                                    & 0.852                                                    & 0.8529 / 0.8909                                              & 0.9182 / 0.8913                                               & 0.9335                                                   \\ \hline
LoRA$_{R=4}$    & 1.26M / 1.26M            & 0.6800                                                    & 0.9121 / 0.9119                                      & 0.9390                                                    & 0.9011                                                    & 0.7183                                                   & 0.8556                                                   & 0.8922 / 0.9217                                              & 0.9120 / 0.8826                                               & 0.9335                                                   \\ \hdashline
AdaLoRA         & 1.92M / 1.26M            & 0.6631                                                    & 0.9054 / 0.9087                                               & 0.9341                                                    & 0.9035                                                    & 0.5634                                                   & 0.8448                                                   & 0.8799 / 0.9111                                              & 0.9122 / 0.8844                                               & \textbf{0.9552}                                          \\ \hdashline
SoRA            & 1.92M / 1.38M \textasciitilde{} 1.55M      & 0.6554                                                    & 0.9049 / 0.9073                                               & 0.9262                                                    & 0.8894                                                   & 0.6901                                                   & 0.8484                                                   & 0.8946 / 0.9239                                              & 0.8196 / 0.7621                                               & 0.8199                                                   \\ \hdashline
\rowcolor{gray!30} R-LoRA$_{R=4}$    & 1.26M / 1.26M            & 0.6852                                                    & 0.9142 / 0.9143                                               & 0.9414                                                    & 0.9032                                                    & 0.7183                                                   & 0.8628                                                   & \textbf{0.8971 / 0.9245}                                     & \textbf{0.9165 / 0.8892}                                               & 0.9427                                                   \\ \hdashline
\rowcolor{gray!30} RM-LoRA$_{R=4}$         & 1.26M / 1.26M            & \textbf{0.6888}                                           & \textbf{0.9148 / 0.9145}                                               & \textbf{0.9418}                                                    & \textbf{0.9062}                                                    & \textbf{0.7324}                                                   & \textbf{0.8664}                                          & 0.8922 / 0.9217                                              & 0.9125 / 0.8839                                               & 0.9427                                                   \\ \hline
LoRA$_{R=8}$      & 1.92M / 1.92M            & 0.6720                                                    & 0.9128 / 0.9137                                               & 0.9378                                                    & 0.8996                                                    & 0.7183                                                   & 0.8559                                                   & 0.8995 / 0.9279                                              & 0.9164 / 0.8887                                               & 0.9358                                                   \\ \hdashline
AdaLoRA         & 3.25M / 1.92M            & 0.6639                                                    & 0.9116 / 0.9131                                               & 0.9418                                           & 0.9055                                                    & 0.5775                                                   & 0.8406                                                   & 0.8971 / 0.9242                                              & 0.9123 / 0.8848                                               & 0.9483                                                   \\ \hdashline
SoRA            & 3.25M / 2.24M \textasciitilde{} 2.84M      & 0.6651                                                    & 0.9070 / 0.9098                                               & 0.9299                                                    & 0.8946                                                    & 0.6901                                                   & 0.8414                                                   & 0.8970 / 0.9266                                              & 0.8188 / 0.7612                                               & 0.8256                                                   \\ \hdashline
\rowcolor{gray!30} R-LoRA$_{R=8}$    & 1.92M / 1.92M            & \textbf{0.6935}                                                    & 0.9135 / 0.9137                                               & 0.9432                                                    & 0.9042                                                    & 0.7324                                                   & 0.8632                                                   & 0.9069 / 0.9331                                              & \textbf{0.9169 / 0.8903}                                      & 0.9404                                                   \\  \hdashline
\rowcolor{gray!30} RM-LoRA$_{R=8}$         & 1.92M / 1.92M            & 0.6878                                                    & \textbf{0.9149 / 0.9150}                                      & \textbf{0.9436}                                           & \textbf{0.9069}                                           & \textbf{0.7465}                                           & \textbf{0.8695}                                                   & \textbf{0.9142 / 0.9378}                                     & 0.9132 / 0.8847                                               & \textbf{0.9483}                                                   \\ \hline
\end{tabular}
\end{center}
\vspace{-0.15cm}
\end{table*}

\begin{figure}
    \centering
    \includegraphics[width=0.45\textwidth]{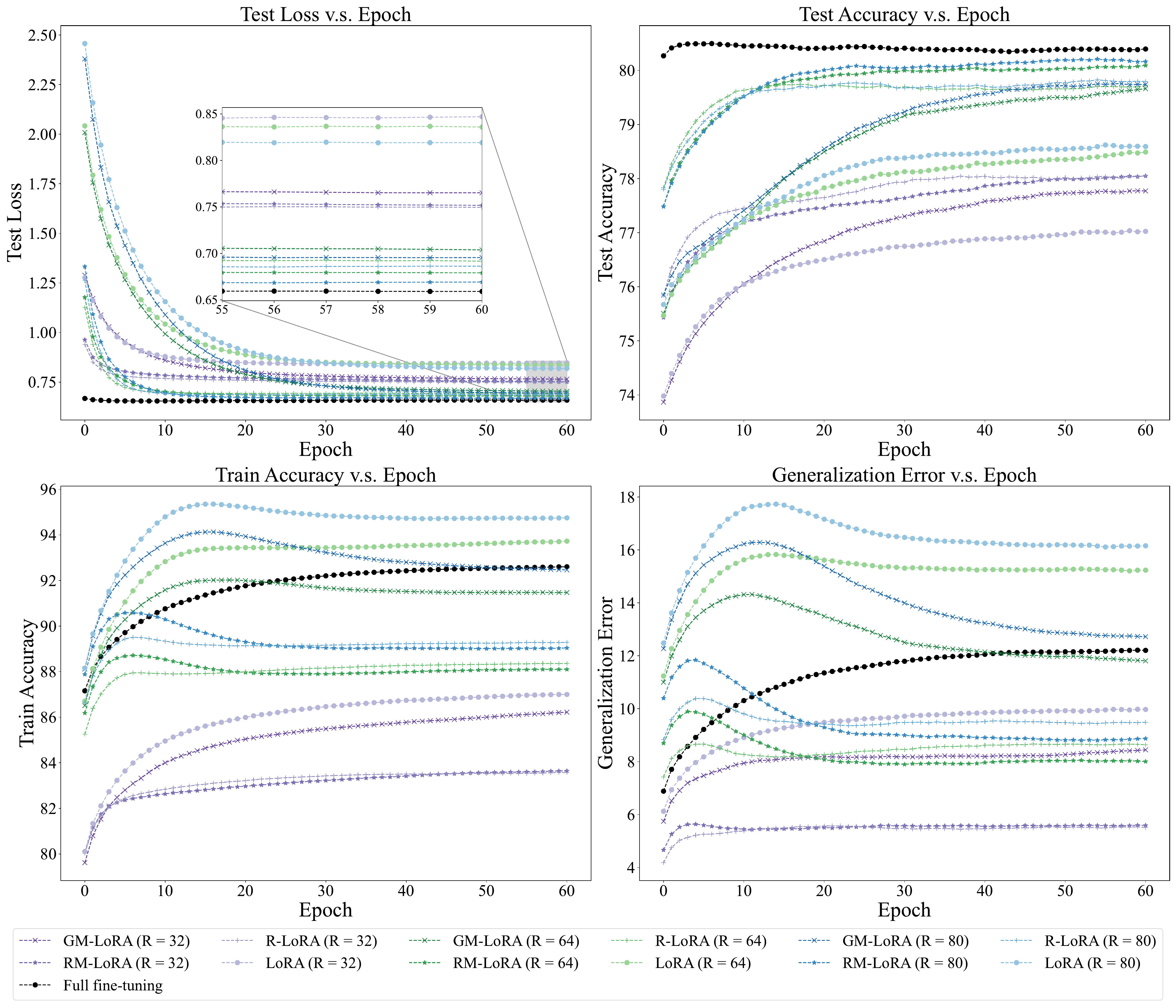}
    \caption{Results with ViT model on CIFAR-100.}
    \label{fig:cifar100_result}
\vspace{-0.35cm}
\end{figure}

\begin{table}
\begin{center}
\caption{Results with DeBERTaV3 model on SQuAD.}
\vspace{-0.25cm}
\renewcommand{\arraystretch}{1.5}
\label{tab:squad_results}
\scriptsize
\begin{tabular}{c|cc|cc|c|c}
\hline
\textbf{Method} & \multicolumn{2}{c|}{\textbf{\# T / I-Rank}} & \multicolumn{2}{c|}{\textbf{\# T / I-Params}} & \textbf{EM} & \textbf{F1 Score} \\ \hline
Full FT & N/A & N/A & \textbf{184M} & \textbf{184M} & \textbf{87.8619} & \textbf{93.6511} \\ \hline
LoRA$_{R=8}$ & 8 & 8 & 1.33M & 1.33M & 87.5118 & 93.3285 \\ \hdashline
AdaLoRA & 16 & 8 & 2.66M & 1.33M & 86.5468 & 92.8831 \\ \hline
SoRA$_{\lambda_2=5e-4}$ & 16 & 16 & 2.66M & 1.28M & 82.7530 & 90.4985 \\ \hdashline
\rowcolor{gray!30} \textbf{RM-LoRA}$_{R=8}$ & 8 & 8 & \textbf{1.33M} & \textbf{1.33M} & \textbf{87.8997} & \textbf{93.6738} \\ \hline
LoRA$_R=4$ & 4 & 4 & 0.67M & 0.67M & 87.4456 & 93.3515 \\ \hdashline
AdaLoRA & 8 & 4 & 1.33M & 0.67M & 86.4995 & 92.8298 \\ \hdashline
SoRA$_{\lambda_2=5e-4}$ & 8 & 8 & 1.33M & 0.61M & 79.6594 & 88.0300 \\ \hdashline
\rowcolor{gray!30} \textbf{RM-LoRA}$_{R=4}$ & 4 & 4 & \textbf{0.67M} & \textbf{0.67M} & \textbf{87.5970} & \textbf{93.5080} \\ \hline
\end{tabular}
\end{center}
\vspace{-0.15cm}
\end{table}

\begin{table}
\begin{center}
\caption{Hyperparameters for GLUE tasks.}
\vspace{-0.25cm}
\renewcommand{\arraystretch}{1.5}
\label{tab:glue_information}
\scriptsize
\begin{tabular}{c|cc|c|c|cc}
    \hline
    \multirow{2}{*}{\textbf{Datasets}} & \multicolumn{2}{c|}{\textbf{Learning Rate}} & \multirow{2}{*}{\textbf{Batch Size}} & \multirow{2}{*}{\textbf{\# Epochs}} & \multicolumn{2}{c}{\textbf{SoRA Parameters}}  \\ \cline{2-3} \cline{6-7}
    & \textbf{FT} & \textbf{LoRA} & & & \textbf{$\lambda$} & \textbf{Sparsity} \\ \hline
    CoLA & 0.00001 & 0.001 & 32 & 3 & 0.001 & 68.96\% \\ \hdashline
    SST-2 & 0.00001 & 0.001 & 32 & 3 & 0.005 & 53.90\% \\ \hdashline
    MRPC & 0.00001 & 0.001 & 32 & 5 & 0.001 & 79.27\% \\ \hdashline
    STS-B & 0.00001 & 0.001 & 32 & 3 & 0.001 & 80.62\% \\ \hdashline
    QQP & 0.00001 & 0.001 & 32 & 3 & 0.005 & 53.90\% \\ \hdashline
    QNLI & 0.00001 & 0.001 & 32 & 3 & 0.0001 & 58.06\% \\ \hdashline
    MNLI & 0.00001 & 0.001 & 32 & 3 & 0.005 & 64.77\% \\ \hdashline
    WNLI & 0.00001 & 0.001 & 32 & 5 & 0.01 & 62.53\% \\ \hdashline
    RTE & 0.00001 & 0.001 & 32 & 3 & 0.005 & 64.23\% \\
    \hline
\end{tabular}
\end{center}
\vspace{-0.55cm}
\end{table}

\subsection{Image Classification}

Figure~\ref{fig:cifar100_result} illustrates the results achieved by the ViT model on the CIFAR-100 dataset, serving as a preliminary measure of the performance of LoRA and the enhancement techniques for LoRA method proposed in this paper. To simulate the theoretical results based on the fully connected layer, only the last classification layer of the ViT model is fine-tuned by LoRA low-rank matrices. The four sub-figures of Figure~\ref{fig:cifar100_result} display the test loss, test accuracy, train accuracy, and generalization error (measured by train accuracy minus test accuracy) for each method respectively.

As observed in Figure~\ref{fig:cifar100_result}, the regularization and gradient masking techniques proposed in this paper both effectively mitigate overfitting and achieve higher accuracy on the test dataset. Specifically, Regularized LoRA (\textbf{R-LoRA}) and Gradient Masking LoRA (\textbf{GM-LoRA}) represent the application of each technique individually, while RM-LoRA combines them together as described in Section~\ref{sec:method}. Furthermore, Table~\ref{table:cifar100_analysis} presents the orthogonal loss $\| \Delta 
 \mathbf{W} (\Delta  \mathbf{W})^{\top} - \mathbf{I} \|_F^2$ of $\Delta \mathbf{W}$ after being fine-tuned by each method, which describes its spatial distribution. The results in Table~\ref{table:cifar100_analysis} demonstrate that the orthogonal penalty term for the LoRA matrices $\mathbf{W}_A$ and $\mathbf{W}_B$ in Eq.~\ref{eq:regularization} effectively promotes the orthogonality of their product. Meanwhile, as the orthogonal loss of $\Delta \mathbf{W}$ decreases, the accuracy of LoRA fine-tuning on the test data increases. Therefore, by effectively promoting the reduction of $\Delta \mathbf{W}$'s orthogonal loss, the RM-LoRA method proposed in this paper achieves the best generalization performance across all rank settings.

\begin{table}
\begin{center}
\caption{Correlation between the orthogonality of $\Delta \mathbf{W}$ and generalization performance.}
\vspace{-0.25cm}
\label{table:cifar100_analysis}
\scriptsize
\renewcommand{\arraystretch}{1.5} 
\begin{tabular}{c|c|c|c}
\hline
\textbf{Method}                    & \multicolumn{1}{c|}{\textbf{Rank of $\Delta \mathbf{W}$}} & \multicolumn{1}{c|}{\textbf{Orthogonality Loss of $\Delta \mathbf{W}$}} & \textbf{Test Acc}         \\ \hline
LoRA$_{R=32}$               & 29                                      & 585.835                                               & 77.024           \\ \hdashline
GM-LoRA$_{R=32}$  & 29                                      & 277.919                                               & 77.772           \\ \hdashline
R-LoRA$_{R=32}$             & 30                                      & 122.079                                               & 78.050           \\ \hdashline
\rowcolor{gray!30} \textbf{RM-LoRA$_{R=32}$}            & 30                                      & \textbf{95.240}                                       & \textbf{78.054}  \\ \hline
LoRA$_{R=64}$               & 55                                      & 410.154                                               & 78.491           \\ \hdashline
GM-LoRA$_{R=64}$  & 52                                      & 175.010                                               & 79.662           \\ \hdashline
R-LoRA$_{R=64}$             & 59                                      & 79.876                                                & 79.721           \\ \hdashline
\rowcolor{gray!30} \textbf{RM-LoRA$_{R=64}$}            & 57                                      & \textbf{58.010}                                       & \textbf{80.092}  \\ \hline
LoRA$_{R=80}$               & 64                                      & 342.197                                               & 78.190           \\ \hdashline
GM-LoRA$_{R=80}$  & 59                                      & 155.010                                               & 79.730           \\ \hdashline
R-LoRA$_{R=80}$             & 72                                      & 68.970                                                & 79.680           \\ \hdashline
\rowcolor{gray!30} \textbf{RM-LoRA$_{R=80}$}            & 69                                      & \textbf{48.832}                                       & \textbf{80.210}  \\ \hline
\end{tabular}
\end{center}
\vspace{-0.35cm}
\end{table}

\subsection{Natural Language Understanding}

The GLUE benchmark includes two single-sentence classification tasks (CoLA, SST-2), three similarity and paraphrase tasks (MRPC, STS-B, QQP), and four natural language inference tasks (QNLI, WNLI, MNLI, RTE). The proposed RM-LoRA method is compared against the baseline methods under multiple LoRA rank settings to demonstrate its superiority. Table~\ref{tab:glue results} shows the performance achieved by different methods on GLUE tasks, as well as the number of trainable and inference parameters (\textit{\# T~/~I~-~Params} respectively). The best result for each task is highlighted in \textbf{bold}. R-LoRA with the proposed regularizer consistently achieves performance gains under the same or lower inference parameter budget compared to other methods in most cases. Furthermore, RM-LoRA with gradient masking outperforms R-LoRA in a majority of task settings. The specific fine-tuning hyperparameters adopted by each method on the GLUE benchmark are summarized in Table~\ref{tab:glue_information}.

\subsection{Question Answering}
The performance of different methods using DeBERTaV3 model on the SQuAD dataset are shown in Table~\ref{tab:squad_results}. Similarly, the proposed RM-LoRA method outperforms other baselines under the same or lower inference parameter budget across varying LoRA rank settings, with EM denoting the average exact match score and F1 referring to the average F1 score. These results highlight that the RM-LoRA method consistently improves LoRA's fine-tuning performance across various benchmarks. The capability of achieving better or comparable performance with reduced parameter budget is especially significant for practical deployment in mobile systems, where efficiency and resource utilization are crucial factors.

\section{Conclusion}

In conclusion, the exploration of intrinsic dimension in LoRA fine-tuning reveals critical insights into optimizing parameter efficiency and enhancing model generalization. The theoretical foundation indicates that the intrinsic dimension of the approximated matrix updates is more pivotal in achieving effective LoRA fine-tuning than the previously emphasized LoRA rank. By employing a regularization technique and a gradient masking method to encourage parameter space exploration while controlling the trainable parameters budget, this paper presents an advanced low-rank adaptation strategy that addresses the challenges of sub-optimal performance and overfitting associated with LoRA. The better generalization performance achieved by the proposed RM-LoRA under the same or lower parameter budget compared to other methods represents significant progress in the field of parameter-efficient fine-tuning for large pre-trained models.

\clearpage
\bibliographystyle{unsrtnat}
\bibliography{sample-acmcp}

\end{document}